\documentclass[letterpaper, 10 pt, conference]{URL-ieeeconf}
\IEEEoverridecommandlockouts    
\usepackage{graphics}           
\usepackage{times}              
\usepackage{amsmath}            
\usepackage{amssymb}            
\usepackage{graphicx}
\usepackage{algorithm}
\usepackage[noend]{algpseudocode}
\usepackage{booktabs}
\usepackage{color}
\usepackage{multirow}
\usepackage{subcaption}
\usepackage{rotating}
\usepackage{cite}
\definecolor{instructioncolor}{rgb}{.5,.5,.5}
\usepackage{tabularx}
\usepackage{booktabs}
\usepackage[table]{xcolor}
\usepackage{array}
\usepackage{siunitx}
\usepackage{caption}
\usepackage{hyperref}
\hypersetup{colorlinks=true, urlcolor=blue, citecolor=black, linkcolor=black}
\let\labelindent\relax
\usepackage{enumitem}
\usepackage{paracol}

\def\eqref#1{(\ref{#1})}

\def\vsfig{\vspace{-0.3cm}}
\def\vstab{\vspace{-0.3cm}}

\newcommand{\rom}[1]{\uppercase\expandafter{\romannumeral #1\relax}}

\makeatletter
\usepackage{xspace}
\DeclareRobustCommand\onedot{\futurelet\@let@token\@onedot}
\def\@onedot{\ifx\@let@token.\else.\null\fi\xspace}

\makeatother

\usepackage{array}
\newcolumntype{L}[1]{>{\raggedright\let\newline\\\arraybackslash\hspace{0pt}}m{#1}}
\newcolumntype{C}[1]{>{\centering\let\newline\\\arraybackslash\hspace{0pt}}m{#1}}
\newcolumntype{R}[1]{>{\raggedleft\let\newline\\\arraybackslash\hspace{0pt}}m{#1}}

\usepackage{makecell}
\usepackage{dsfont}
\usepackage{siunitx}
\usepackage{threeparttable}
\usepackage{tabularx}
\usepackage{supertabular}
\usepackage[table]{xcolor} 
\usepackage{booktabs}
\usepackage{amssymb}
\definecolor{c1st}{RGB}{140,214,164}
\definecolor{c2nd}{RGB}{209,237,217}
\definecolor{ourrow}{RGB}{246,246,246}
\newcommand{\fst}[1]{\cellcolor{c1st}\textbf{#1}}
\newcommand{\snd}[1]{\cellcolor{c2nd}#1}
\newcommand{\up}{\textcolor{black}{\scriptsize$\uparrow$}}
\newcommand{\dn}{\textcolor{black}{\scriptsize$\downarrow$}}
\newcommand{\pz}{\phantom{0}}

\newcommand{\skatenav}{SkateNav}
\hypersetup{citecolor=blue, linkcolor=blue} 
\title{\vspace{-24pt}\LARGE \bf SkateNav: Learning Energy-Efficient Navigation through \\ Adaptive Skating on Passive-Wheel Quadrupedal Robots}
\author{%
  Jeil~Jeong$^{*1,2}$,
  Taeyun~Kim$^{*1}$,
  Sangjin~Kim$^{1}$,
  Jinyoung~Choi$^{1}$,
  and Sung-Eui~Yoon$^{\dagger2}$%
  \thanks{\scriptsize $^{*}$Equal contribution. $^{\dagger}$Corresponding author: sungeui@kaist.edu}%
  \thanks{\scriptsize $^{1}$NAVER LABS, Republic of Korea.
    \{ktw1404,s.j.kim,jy-choi\}@naverlabs.com}%
  \thanks{\scriptsize $^{2}$School of Computing, KAIST, Republic of Korea.
    jeil.jeong@kaist.ac.kr}%
}

\begin{document}
\maketitle
\thispagestyle{empty}
\pagestyle{empty}

\begin{abstract}
Passive-wheel quadrupedal robots offer energy-efficient locomotion through leg-driven propulsion and momentum-driven gliding.
However, navigation requires coordinating steering with intermittent propulsion and adapting push--glide timing to task demands.
Velocity-tracking objectives may suppress the speed variations inherent in skating, discouraging efficient gliding.
To address these challenges, we propose SkateNav, a unified skating-navigation framework based on reinforcement learning.
A time-budgeted goal-reaching objective allows the policy to jointly learn steering, propulsion, and their temporal allocation.
An adaptive skating cycle structures gliding, pushing, and recovery, while frequency modulation at every control step adjusts phase durations to navigation demands.
Simulation experiments demonstrate higher average success rates and lower cost of transport than the evaluated planner--controller baselines, with ablations confirming the contributions of cycle structure and frequency adaptation.
Real-world experiments demonstrate zero-shot transfer from simulation to a passive-wheel Mini Cheetah across cluttered indoor spaces, slopes, and paved outdoor surfaces.
Supplementary videos are available on the \href{https://ktw1404.github.io/SkateNav/}{project page}.
\end{abstract}

\section{Introduction}
\label{sec:intro}
Quadrupedal robots can adapt their footholds and posture to traverse rough and discontinuous terrain~\cite{miki2022learning, agarwal2023legged, cheng2024extreme}.
However, they generally have lower energy efficiency than wheeled platforms on flat terrain, partly due to energy loss during foot--ground impacts~\cite{seok2014design}.
To address this, wheeled-legged robots have been proposed to combine the terrain adaptability of legs with the energy efficiency of wheels.
These systems can be classified into active-wheel configurations with dedicated wheel actuators for propulsion and braking~\cite{bjelonic2021whole, lee2024learning} and passive-wheel configurations without additional wheel actuators~\cite{chen2022trot}.
Passive-wheel systems use leg-driven pushes to build momentum and exploit low rolling resistance for energy-efficient gliding, as in human skating~\cite{de1989gliding}.

Model-based controllers realize skating by coordinating leg motion and contact forces under nonholonomic rolling constraints~\cite{bjelonic2018skating, chen2023locomotion}.
Recent reinforcement learning (RL) approaches improve speed and robustness but primarily track prescribed velocity commands~\cite{gu2026skater, wang2026efficient}.
Leg-driven pushes provide propulsion and steering at an energy cost, whereas gliding sustains motion with lower energy use but offers limited control over propulsion and steering.
Energy-efficient navigation therefore requires deciding when to push and how long to glide to balance timely goal arrival and energy use.
However, velocity-tracking objectives may penalize the speed variations inherent in push--glide motion, discouraging sustained gliding and limiting the allocation of pushing and gliding according to navigation demands.
Thus, a planner and controller interfaced through velocity commands may not fully accommodate such push--glide timing. 

A goal-reaching objective specifies a destination and time budget~\cite{rudin2022advanced}, allowing the velocity profile to vary.
This flexibility alone, however, does not ensure that efficient push--glide behavior emerges.
Predefined skating cycles provide motion structure~\cite{bjelonic2018skating}, but fixed contact timings limit adaptation to navigation demands.
The challenge is therefore to preserve the skating structure while adapting when to push and how long to glide for energy-efficient navigation.

\begin{figure}[t!]
    \captionsetup{font=footnotesize}
    \centering
    \includegraphics[width=\columnwidth]{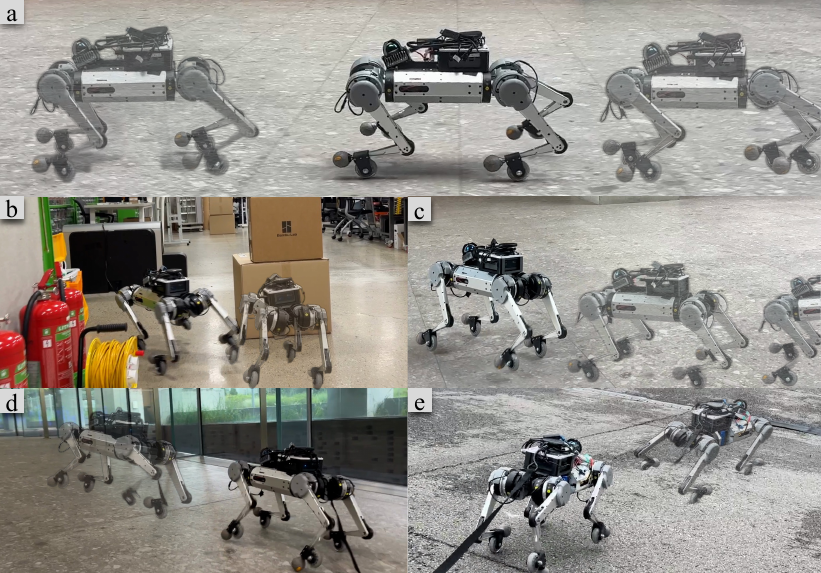}
    \caption{
    \textbf{SkateNav for passive-wheel navigation in diverse environments.}\linebreak
    Our SkateNav framework enables local navigation of a Mini Cheetah~\cite{katz2019mini} equipped with custom detachable passive-wheel modules.
    Representative snapshots show the robot navigating toward the goal through (a) gliding on passive wheels, (b) obstacle avoidance in a cluttered passage, (c) heading adjustment, and (d, e) uphill and downhill slope traversal.}
    \label{fig:overview}
    \vspace{-20pt}
\end{figure}

To address this, we propose SkateNav, a unified skating-navigation policy for energy-efficient local navigation of passive-wheel quadrupedal robots (Fig.~\ref{fig:overview}).
Given waypoints, remaining time, and terrain and robot observations, the policy jointly determines steering, propulsion, and skating-cycle progression.
The cycle organizes skating motion into gliding, pushing, and leg recovery in the phase domain, while the policy modulates its frequency at every control step to adapt phase durations to navigation demands.

The main contributions of this work are:
\begin{itemize}[leftmargin=*, topsep=2pt, itemsep=1pt, parsep=0pt]
    \item We formulate passive-wheel local navigation as a time-budgeted goal-reaching problem in which steering, propulsion, and push--glide timing are learned jointly.
    \item We introduce an adaptive skating cycle representation that preserves leg coordination while exposing phase progression as a navigation-conditioned policy action.
    \item We demonstrate higher average success rates and lower cost of transport than the evaluated planner--controller baselines, conduct component ablations, and validate real-world navigation across indoor and outdoor environments.
\end{itemize}

\section{Related Work} 
\label{sec:related}
\subsection{Passive-Wheel Legged Robot Locomotion}
Passive-wheel legged robots achieve energy efficiency by generating propulsion through leg-driven ground pushes and gliding on the resulting momentum~\cite{endo2011study, chen2022trot}. 
Prior studies have explored control strategies for passive-wheel skating behavior~\cite{hashimoto2008swizzle,kalouche2015modularity,geilinger2020computational}.
Model-based controllers coordinate body motion and contact forces under nonholonomic constraints to realize designed skating gaits~\cite{bjelonic2018skating,chen2023locomotion}.
Recently, RL approaches have improved skating speed and robustness by leveraging contact modeling, constraint-aware rewards and hardware-policy co-design~\cite{marot2026reinforcement, gu2026skater, wang2026efficient}.
While prior studies focus on realizing robust skating motion, they do not address how skating behavior should adapt to navigation demands.
Passive-wheel navigation requires steering to be coordinated with intermittent leg-generated propulsion and retained rolling momentum.
Moreover, energy-efficient gliding must be balanced with sufficient pushing for timely task completion.
To this end, our SkateNav framework learns a unified skating-navigation policy that jointly coordinates steering and propulsion under a goal-reaching objective while determining the durations of pushing and gliding as part of the navigation decision.

\subsection{Gait Encoding for Skating Motion}
In RL-based legged locomotion, gait patterns have been discovered without explicit structure~\cite{hwangbo2019learning} or under only implicit guidance penalizing excessive swing times~\cite{rudin2022learning, ji2022concurrent}.
To achieve diverse and natural gaits, however, preferred patterns have been structured through phase-based encodings such as central pattern generators~\cite{lee2020learning, miki2022learning}.
For passive-wheel systems, such structure can help separate propulsion from gliding, as learned policies tend to maintain wheel--ground contact without clearly distinguishing these phases~\cite{wang2026efficient}.
Accordingly, prior work has encoded human-inspired skating gaits as time-phased push--glide cycles, in which the legs alternate with a phase offset~\cite{bjelonic2018skating}.
However, these cycles use predefined contact timings, making it difficult to accommodate varying navigation demands, from repeated pushes for acceleration or rapid steering for obstacle avoidance to prolonged gliding when momentum suffices.
We therefore let the policy modulate the cycle frequency online, adapting the durations of pushing and gliding according to task requirements.

\subsection{Goal-Conditioned Local Navigation}
Recent RL approaches train a single goal-conditioned policy that directly maps a local waypoint and sensory observations to joint-level actions~\cite{zhuang2023robot, cheng2024extreme, yang2022learning}.
A goal-reaching objective defined by a time budget rather than instantaneous velocity commands allows policies to discover embodiment--specific behaviors such as terrain-negotiating gaits~\cite{rudin2022advanced} or high-speed obstacle avoidance~\cite{AgileButSafe}.
Existing learning-based skating controllers have primarily been trained to track prescribed velocity commands~\cite{marot2026reinforcement, gu2026skater, wang2026efficient}.
For passive-wheel systems, such objectives may suppress the speed variations inherent in the push--glide cycle and discourage energy-efficient gliding.
Accordingly, SkateNav adopts a goal-conditioned formulation, providing temporal flexibility for the policy to shape its motion according to passive-wheel dynamics and task demands rather than prescribed instantaneous velocities.

\section{Mechanical Design} \label{sec:mechanical design}
\vspace{-2mm}
\begin{figure}[h!]
	\captionsetup{font=footnotesize}
	\centering
	\includegraphics[width=\columnwidth]{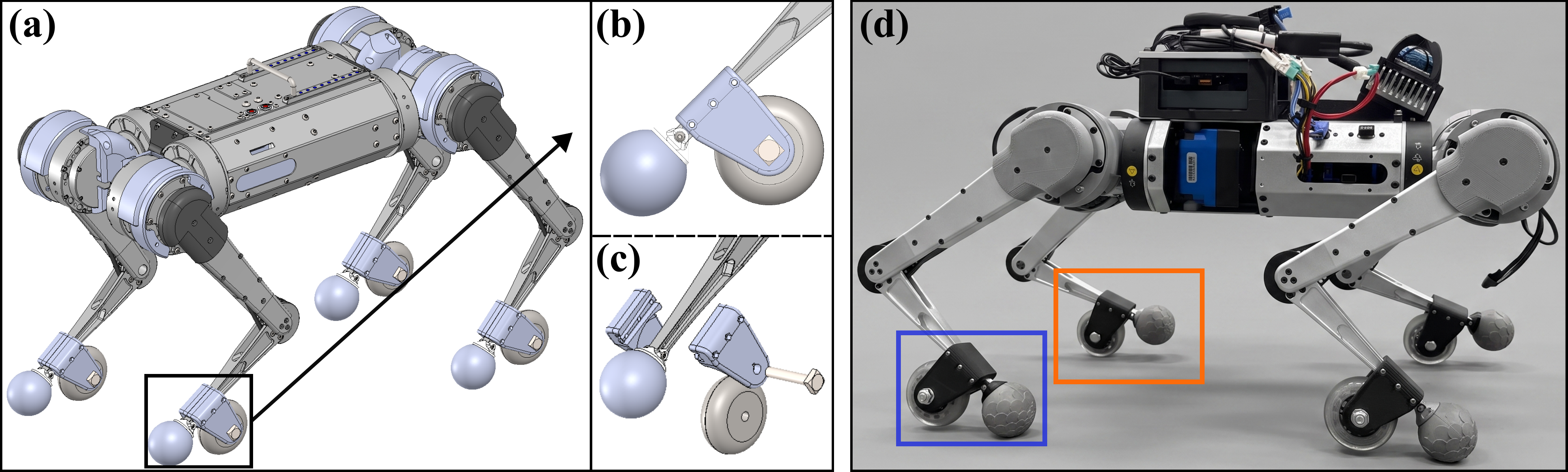}
	\caption{
    \textbf{Mechanical design and hardware implementation.} 
    (a) Overall design of the quadruped robot equipped with passive-wheel modules. 
    (b,c) Side and assembly views of the foot and detachable passive-wheel module.
    (d) Hardware implementation, with the blue and orange boxes indicating foot-ground and wheel-ground contact modes, respectively.}
    \label{fig:mechanical_design}
    \vsfig
\end{figure}
Human inline skaters exploit the anisotropic rolling constraint of the skate, orienting it relative to the heading direction and pushing laterally to convert the ground reaction force into forward propulsion~\cite{chen2019motion, de1989gliding}.
Reproducing this mechanism on a quadrupedal robot, e.g., Mini Cheetah~\cite{katz2019mini}, requires independent control of the end-effector orientation, but the three actuated degrees of freedom (DoFs) per leg control only its position.
This makes it difficult to generate such propulsion through wheel-ground interaction alone, and prior work has therefore introduced an additional non-rolling contact feature alongside the passive wheel to provide traction for propulsion and braking~\cite{bjelonic2018skating, chen2023locomotion}.
Following this approach, we use the existing foot as the non-rolling contact and attach a detachable passive-wheel module, as shown in Fig.~\ref{fig:mechanical_design}.
This configuration provides a mechanical basis for skating motion: foot-ground contact generates propulsion, whereas wheel-ground contact provides low-resistance rolling.
The skating-navigation policy exploits these complementary contact modes to adaptively coordinate pushing and gliding duration for energy-efficient navigation.

\section{Variable Notation}
\label{sec:variable}
We denote Cartesian position, linear velocity, and linear acceleration by $\boldsymbol{p}$, $\boldsymbol{v}$, and $\dot{\boldsymbol{v}} \in \mathbb{R}^3$, respectively.
Orientation is parameterized by the XYZ Euler angles $\boldsymbol{\theta} \in \mathbb{R}^3$, and angular velocity is denoted by $\boldsymbol{\omega} \in \mathbb{R}^3$.
The robot has twelve actuated joints, whose positions, velocities, accelerations, and actuation torques are denoted by $\boldsymbol{q}$, $\dot{\boldsymbol{q}}$, $\ddot{\boldsymbol{q}}$, and $\boldsymbol{\tau} \in \mathbb{R}^{12}$, respectively, and four passive-wheel joints, whose positions and velocities are denoted by $\boldsymbol{\psi}$ and $\dot{\boldsymbol{\psi}} \in \mathbb{R}^{4}$, respectively.
The contact states of foot $F_i$ and wheel $W_i$ are represented by $I_{F_i}, I_{W_i} \in \{0,1\}$, where $i \in \{0,1,2,3\}$ indexes the legs in the order Front Right (FR), Front Left (FL), Rear Right (RR), and Rear Left (RL).
Superscripts denote reference frames (world $\mathcal{W}$ or body $\mathcal{B}$), whereas subscripts denote entities, components, or time indices.
For example, $v^{\mathcal{W}}_{B,z,t}$ denotes the $z$-component of the body linear velocity in the world frame at time $t$.
Component and time indices are omitted when clear from context.









\begin{figure*}[ht]
    \captionsetup{font=footnotesize}
    \centering
    \includegraphics[width=\textwidth]{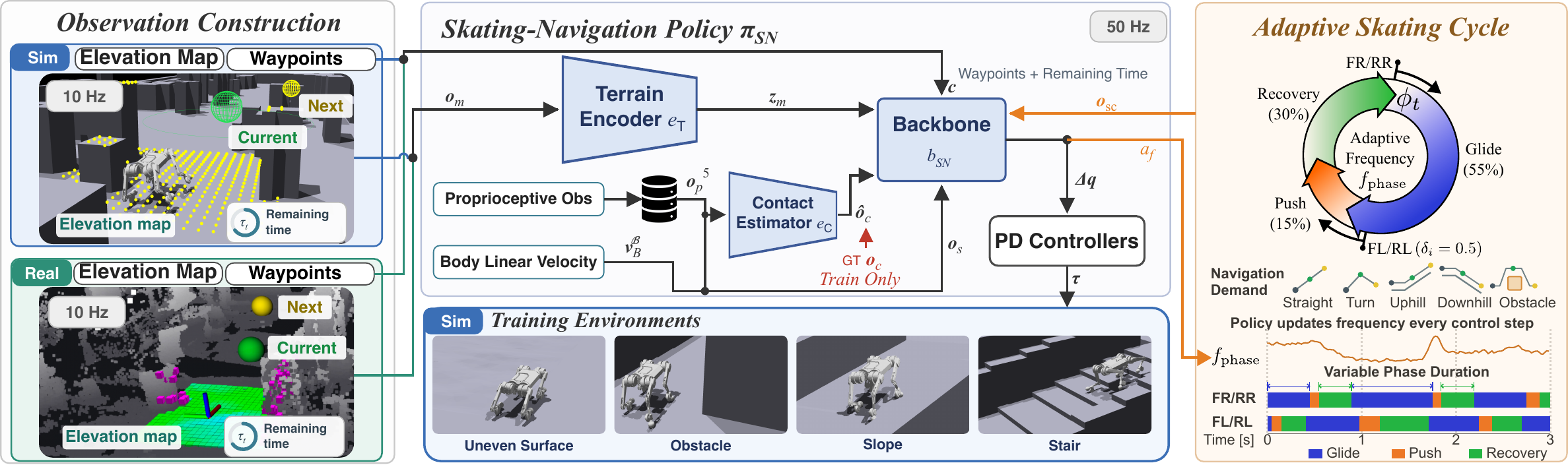}
    \caption{
    \textbf{Overview of the proposed SkateNav framework.}
    A unified skating-navigation policy uses a 2.5D elevation map, robot states, and navigation commands to generate joint actions and a frequency action.
    The navigation commands specify the current and next waypoints and the time remaining to reach the current waypoint.
    The frequency action controls progression of the skating cycle, allowing phase durations to vary with navigation demands.
    }
    \label{fig:framework}
    \vsfig
\end{figure*}

\section{Methodology}
\subsection{Characteristic of Passive-Wheel System}
\vspace{-4mm}
\begin{figure}[h!]
    \captionsetup{font=footnotesize}
    \centering
    \includegraphics[width=\columnwidth]{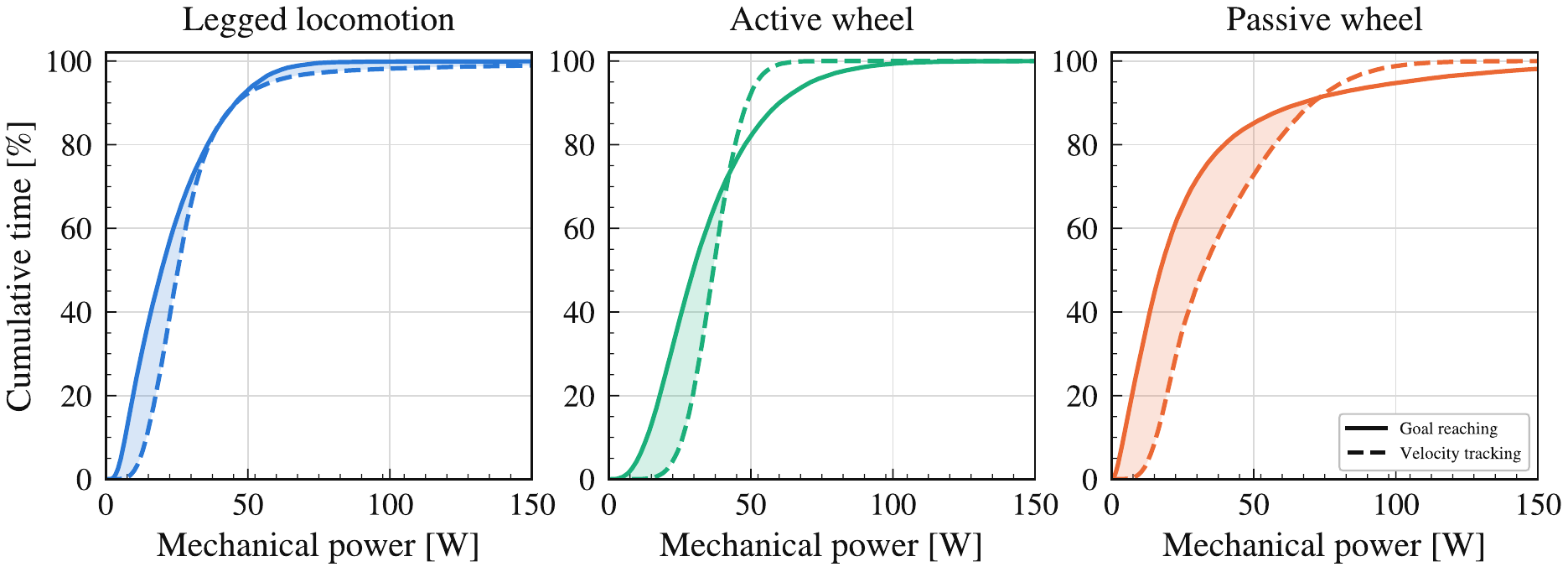}
    \vspace{-6mm}
    \caption{
    Each panel shows the cumulative distributions of instantaneous mechanical power for policies trained with goal-reaching (solid) and velocity-tracking (dashed) objectives on each platform, averaged over 100 trials.
    Both policies traverse \SI{7}{\meter} in \SI{7}{\second}, with a \SI{1}{\meter/\second} command for velocity tracking.
    The vertical axis shows the percentage of time at or below each power level; shading highlights the difference in the low-power regime.
    }
    \label{fig:design_choice}
    \vsfig
\end{figure}
Legged and actively wheeled systems can continuously generate propulsion through direct ground interaction.
In contrast, passive-wheel systems lack wheel-driven propulsion and therefore rely on intermittent leg-driven pushes followed by gliding on the resulting momentum.
Push--glide motion inherently involves speed variations, and the extent to which these variations are tolerated under the control policy's learning objective can affect push--glide timing and power use.
A velocity-tracking objective penalizes deviations from a prescribed instantaneous velocity and may therefore discourage these variations, whereas a goal-reaching objective specifies a destination and time budget, providing temporal flexibility, allowing the policy to adjust the velocity profile to accommodate push--glide motion.
Fig.~\ref{fig:design_choice} illustrates how this objective choice affects instantaneous mechanical power distributions.
This redistribution is more noticeable for passive wheels, where goal-reaching increases the time spent in both low- and high-power regimes relative to the velocity-tracking objective.

We aim to exploit this flexibility for energy-efficient local navigation with passive-wheel quadrupeds.
Beyond this flexibility, the policy must structure push--glide motion, coordinate steering with intermittent foot-ground propulsion, and adapt their timing to navigation demands.
Our SkateNav provides this structure and adaptability through an alternating skating cycle with online frequency modulation.
Subsequent sections detail the task, learning objective, adaptive skating cycle, policy network architecture, and implementation.


\subsection{Task and Environment Formulation}
We formulate the task as navigating a passive-wheel quadruped through a sequence of waypoints $\mathcal{G}=(\boldsymbol{G}_k)_{k=1}^{K}$, with each waypoint $\boldsymbol{G}_k \in \mathbb{R}^3$ to be reached within a prescribed time budget.
The waypoint sequence is provided by a global planner~\cite{hart1968astar} that captures the coarse structure of the environment, including diverse terrain geometries such as slopes, stairs, and uneven surfaces, while local obstacles remain unknown a priori.
These variations and changing navigation demands require the policy to adapt its skating behavior and reactively avoid obstacles while operating under nonholonomic rolling constraints.

\subsection{Policy Learning Objective}
The skating-navigation policy $\pi_{\mathrm{SN}} : \mathcal{O} \times \mathcal{C} \rightarrow \mathcal{A}$ is optimized via reinforcement learning (RL) under a partially observable Markov decision process (POMDP), defined by the tuple $(\mathcal{S}, \mathcal{O}, \mathcal{A}, \mathcal{P}, \mathcal{R}, \rho_0, \gamma)$, which specifies the state, observation, and action spaces, transition dynamics, reward function, initial-state distribution, and discount factor, respectively.
The navigation command $\boldsymbol{c}_t = [\boldsymbol{G}^{\mathcal{B}}_{\mathrm{cur},t}, \boldsymbol{G}^{\mathcal{B}}_{\mathrm{nxt},t}, \tau_t] \in \mathcal{C}$ consists of the current and subsequent waypoints, together with the normalized remaining time $\tau_t \in [0,1]$ for the current waypoint.
The subsequent waypoint provides route look-ahead and coincides with the current one at the final goal.


At the beginning of each episode, the robot is initialized with its nominal standing posture $\boldsymbol{q}_0$, with the initial state sampled as $\boldsymbol{s}_0 \sim \rho_0$.
We optimize the skating-navigation policy parameters $\phi_{\mathrm{SN}}$ to maximize the expected return over the distribution of navigation commands $P(\boldsymbol{c})$:
\begin{equation}
\pi_{\mathrm{SN}} := \pi_{\mathrm{SN}}(\cdot;\phi_{\mathrm{SN}}^{*}), 
\quad 
\phi_{\mathrm{SN}}^{*} = \arg\max_{\phi_{\mathrm{SN}}} J(\phi_{\mathrm{SN}}),
\end{equation}
\vspace{-2mm}
\begin{equation}
\resizebox{0.90\columnwidth}{!}{$\displaystyle
J(\phi_{\mathrm{SN}}) = 
\mathbb{E}_{\boldsymbol{c} \sim P(\boldsymbol{c})}
\left[
\mathbb{E}_{\substack{
(\boldsymbol{s}, \boldsymbol{a}) \sim \rho_{\phi_{\mathrm{SN}}} \\
\boldsymbol{s}_0 \sim \rho_0}}
\left[
\sum_{t=0}^{\infty}
\gamma^{t}
\mathcal{R}
\left(\boldsymbol{s}_t,\boldsymbol{a}_t \mid \boldsymbol{c}_t\right)
\right]
\right],
$}
\label{eq:objective_function}
\end{equation}
where $\rho_{\phi_{\mathrm{SN}}}$ denotes the state-action visitation distribution under the skating-navigation policy.
\clearpage

\subsection{Adaptive Skating Cycle}
\begin{figure}[h]
    \vspace{-2mm}
	\captionsetup{font=footnotesize}
	\centering
	\includegraphics[width=\columnwidth]{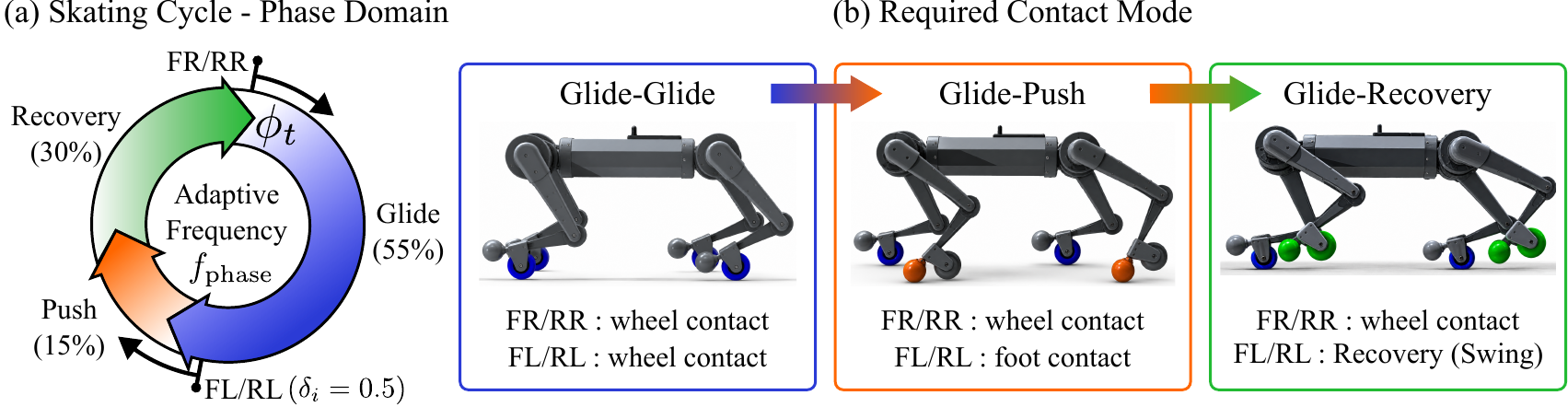}
	\caption{
    (a) Each leg sequentially follows Glide, Push, and Recovery in the phase domain, with the right and left leg pairs offset by half a cycle.
    (b) As the cycle progresses, the desired contact configurations are defined: while the right-side legs remain in Glide, the left-side legs transition through Glide, Push, and Recovery with wheel, foot, and no contact, respectively.
    }
    \label{fig:skating_cycle}
	\vsfig
\end{figure}
Human skating motion is characterized by the alternating roles of the left and right legs~\cite{chen2019motion}. 
While one leg pushes against the ground for propulsion, the other maintains rolling contact and glides on the resulting momentum.
To realize this alternating pattern on a quadruped, we synchronize the front and rear legs on each side and impose a half-cycle phase offset between the left and right pairs.
We represent this coordination using a shared normalized skating phase $\phi_t \in [0,1)$ and a leg-dependent phase offset $\delta_i$, as
\begin{equation} 
\phi_{i,t} = \left( \phi_t + \delta_i \right) \bmod 1, \quad \boldsymbol{\delta} = \left[ 0,\, 0.5,\, 0,\, 0.5 \right]. 
\label{eq:leg_phase} 
\end{equation} 

Within this alternating structure, each leg follows a skating cycle composed of three sequential phases as shown in Fig.~\ref{fig:skating_cycle}(a).
The stage of leg $i$ is determined by its phase as
\begin{equation} 
\Gamma(\phi_{i,t}) =
\begin{cases}
\mathrm{\textit{Glide}}, & 0 \leq \phi_{i,t} < \rho_{\mathrm{G}},\\
\mathrm{\textit{Push}}, & \rho_{\mathrm{G}} \leq \phi_{i,t} < \rho_{\mathrm{G}}+\rho_{\mathrm{P}},\\
\mathrm{\textit{Recovery}}, & \rho_{\mathrm{G}}+\rho_{\mathrm{P}} \leq \phi_{i,t} < 1,
\end{cases}
\label{eq:skating_stage}
\end{equation}
where $\rho_{\mathrm{G}}=0.55$, $\rho_{\mathrm{P}}=0.15$, and $\rho_{\mathrm{R}}=0.30$ are fixed phase-domain fractions, not time-domain proportions.
During the \textit{Glide} phase, only the passive wheel remains in contact to exploit momentum, whereas \textit{Push} phase uses foot-ground contact for propulsion and \textit{Recovery} phase lifts the leg for the next stroke. 
Combined with the half-cycle offset, these stages define the contact configurations in Fig.~\ref{fig:skating_cycle}(b), encouraged by the skating gait reward term in Sec.~\ref{sec:Reward}.
\vspace{-7pt}
\begin{figure}[h!]
	\captionsetup{font=footnotesize}
	\centering
	\includegraphics[width=\columnwidth]{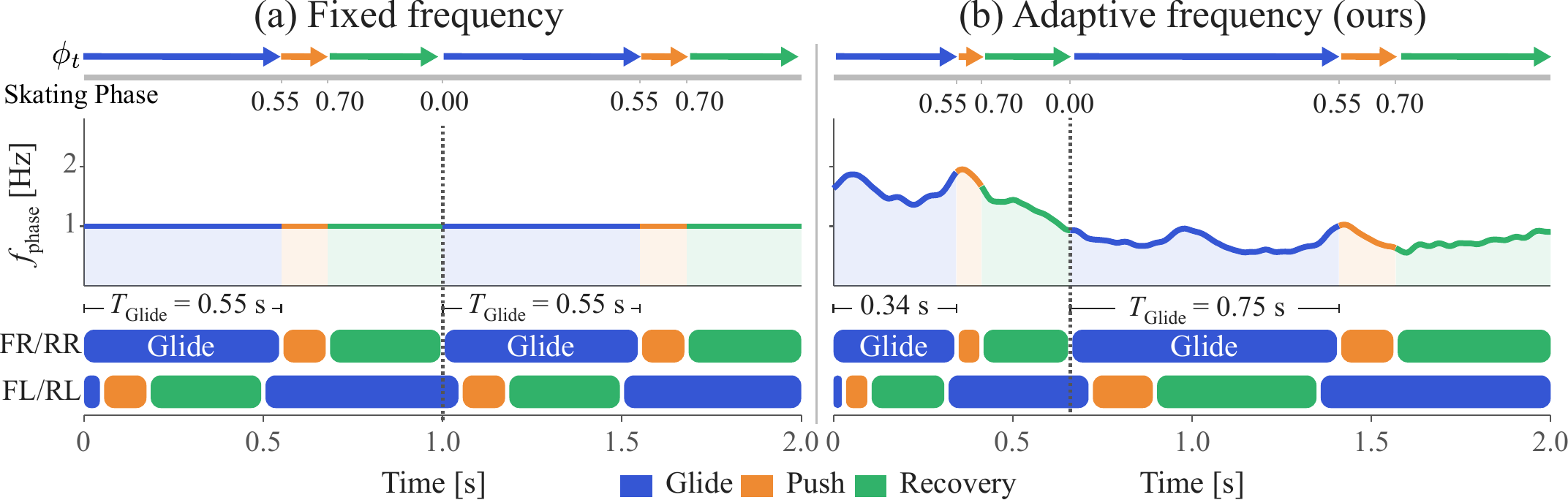}
	\caption{
    (a) A fixed skating frequency advances the phase with predefined durations, while (b) an adaptive skating frequency is adjusted at every control step, allowing durations to vary according to navigation demands.}
    \label{fig:adaptive_frequency}
	\vsfig
\end{figure}

The progression of the shared normalized skating phase over time is determined by the skating frequency $f_{\mathrm{phase},t}$ as
\begin{equation}
\phi_{t+1} = \left(\phi_t + f_{\mathrm{phase},t}\Delta t\right) \bmod 1,
\label{eq:phase_progression}
\end{equation}
where $\Delta t$ denotes the control time step.
With a fixed skating frequency~\cite{bjelonic2018skating}, the predefined phase fractions result in a fixed temporal allocation among \textit{Glide}, \textit{Push}, and \textit{Recovery}, as illustrated in Fig.~\ref{fig:adaptive_frequency}(a). 
Such fixed timing cannot accommodate varying navigation demands, such as more frequent \textit{Push} phases to establish foot-ground contact for rapid steering around obstacles or prolonged \textit{Glide} phases when sufficient momentum is available.
To provide this gait-level flexibility, we allow the policy to modulate $f_{\mathrm{phase},t}$ at every control step according to the navigation context, as shown in Fig.~\ref{fig:adaptive_frequency}(b).
Specifically, the policy outputs a frequency action $a_{f,t}$, mapped to the skating frequency as
\begin{equation}
f_{\mathrm{phase},t} = \operatorname{clip}\left(f_{\mathrm{nom}} + s_f a_{f,t},\, 0.0,\, 3.0\right),
\label{eq:frequency_action_mapping}
\end{equation}
where $f_{\mathrm{nom}}=1.5$ Hz denotes the nominal skating frequency, $s_f=0.25$ is the frequency action scale.
Modulating phase progression within each cycle allows the policy to vary the relative durations of \textit{Glide}, \textit{Push}, and \textit{Recovery} while retaining the phase sequence and inter-leg phase offsets.

\subsection{Policy Architecture}
\setlength{\tabcolsep}{4pt} 
\begin{table}[t!]
\centering
\captionsetup{font=small}
\caption{Domain Randomization Parameters}
\vspace{-2mm}
\begin{tabular}{c|c|c|c}
\hline
\textbf{Parameter} & \textbf{Train Distribution} & \textbf{Test Distribution} & \textbf{Unit} \\
\hline
Payload & $\mathcal{U}(0.0,\,3.0)$ & $\mathcal{U}(0.0,\,5.0)$  & \si{\kilogram} \\
CoM Offset  & $\mathcal{U}(-0.1,\,0.1)^3$  & $\mathcal{U}(-0.15,\,0.15)^3$  & \si{\meter} \\
Wheel Friction & $\mathcal{U}(5\mathrm{e}{-4},\,4\mathrm{e}{-3})$  & $\mathcal{U}(0.0,\,1\mathrm{e}{-2})$  & \si{\newton\meter} \\
Wheel Armature & $\mathcal{U}(2\mathrm{e}{-5},\,1\mathrm{e}{-4})$  & $\mathcal{U}(0.0,\,1\mathrm{e}{-3})$  & \si{\kilogram\meter\squared} \\
Terrain Friction  & $\mathcal{U}(0.7,\,1.2)$  & $\mathcal{U}(0.4,\,1.5)$  & -- \\
Terrain Restitution  & $\mathcal{U}(0.0,\,0.1)$  & $\mathcal{U}(0.0,\,0.3)$  & -- \\
Motor Torque Scale  & $\mathcal{U}(0.9,\,1.1)^{12}$  & $\mathcal{U}(0.8,\,1.3)^{12}$  & -- \\
\hline
\end{tabular}
\begin{tablenotes}
\setlength{\itemindent}{-0.3cm}
\item[] \textbullet\; $\mathcal{U}(a,b)^c$: uniform distribution over $[a,b]$ in each of the $c$ dimensions.
\end{tablenotes}
\vspace{-8mm}
\label{table:dom-rnd}
\end{table}

We employ a concurrent learning framework~\cite{ji2022concurrent} with an asymmetric actor--critic architecture for robust sim-to-real transfer~\cite{nahrendra2023dreamwaq}. 
As shown in Fig.~\ref{fig:framework}, the skating-navigation policy $\pi_{\mathrm{SN}}$ comprises a terrain encoder $e_{\mathrm{T}}$, a contact estimator $e_{\mathrm{C}}$, and a backbone $b_{\mathrm{SN}}$. 
It receives a 5-tuple observation $\boldsymbol{o}=(\boldsymbol{o}_{m}, \boldsymbol{o}_{s}, \boldsymbol{o}_{\mathrm{sc}}, \hat{\boldsymbol{o}}_{c}, \boldsymbol{c}) \in \mathcal{O} \subset \mathcal{S}$,
consisting of
(1) a local elevation map $\boldsymbol{o}_{m}\in\mathbb{R}^{21\times21}$ at \SI{0.1}{\meter} resolution and encoded as $e_{\mathrm{T}}:\boldsymbol{o}_{m}\rightarrow\boldsymbol{z}_{m}\in\mathbb{R}^{32}$; 
(2) a robot state observation $\boldsymbol{o}_{s} = \left[\boldsymbol{o}_{p}^{5}, \boldsymbol{v}_{B}^{\mathcal{B}}\right]$, combining a proprioceptive part $\boldsymbol{o}_{p} = \left[\boldsymbol{q},\,\dot{\boldsymbol{q}},\,\left(\boldsymbol{p}^{\mathcal{B}}_{F_i}\right)_{i},\,\left(\boldsymbol{p}^{\mathcal{B}}_{W_i}\right)_{i},\,\boldsymbol{\theta}^{\mathcal{W}}_{B,xy},\,\boldsymbol{\omega}^{\mathcal{B}}_{B},\,\boldsymbol{a}_{t-1}\right]$, including the joint states, the foot and wheel positions of each leg, the body orientation and angular velocity, and the previous action, with the body linear velocity $\boldsymbol{v}_{B}^{\mathcal{B}}$; 
(3) a skating-cycle observation $\boldsymbol{o}_{\mathrm{sc}} = \left[\left(\sin(2\pi\phi_i)\right)_{i},\,\left(\cos(2\pi\phi_i)\right)_{i}\right]$ that encodes the skating phase of each leg; 
(4) an estimated contact state $\hat{\boldsymbol{o}}_{c}$ predicted by the contact estimator $e_{\mathrm{C}}:\boldsymbol{o}_{p}^{5}\rightarrow\hat{\boldsymbol{o}}_{c}$ from a five-step proprioceptive history, with the ground-truth contact state $\boldsymbol{o}_{c}=\left[\left(I_{F_i}\right)_{i},\,\left(I_{W_i}\right)_{i}\right]$ used to supervise $e_{\mathrm{C}}$ during training; 
and (5) the navigation command $\boldsymbol{c}$. 

The policy backbone $b_{\mathrm{SN}}:\boldsymbol{z}_{m}\times\boldsymbol{o}_{s}\times\boldsymbol{o}_{\mathrm{sc}}\times\hat{\boldsymbol{o}}_{c}\times\boldsymbol{c}\rightarrow\boldsymbol{a}$ outputs $\boldsymbol{a}=\left[\Delta\boldsymbol{q},\,a_{f}\right]\in\mathbb{R}^{13}$, where $\Delta\boldsymbol{q}\in\mathbb{R}^{12}$ specifies the desired joint displacements and $a_{f}$ controls the progression of the skating cycle. 
The joint torques $\boldsymbol{\tau}$ are generated by proportional--derivative controllers to track the targets $\boldsymbol{q}_{0}+\Delta\boldsymbol{q}$.
The asymmetric critic additionally receives the privileged randomized parameters $\boldsymbol{x}_{d}$ summarized in Table~\ref{table:dom-rnd}.

\subsection{Implementation Details}
\label{sec:implementation}
We use the Mini Cheetah robot~\cite{katz2019mini} with custom passive-wheel modules and Isaac Gym simulator~\cite{makoviychuk2021isaac} for large-scale parallel data collection.
Training is conducted on a desktop with an Intel Core i7-13700 CPU and NVIDIA RTX 4090 GPU for 9 hours over 10,000 iterations using 4096 simulated robots.
The skating--navigation policy $\pi_{\mathrm{SN}}$ runs at 50 Hz and is optimized using Proximal Policy Optimization (PPO)~\cite{schulman2017proximal}.
For sim-to-real compatibility, the local elevation map $\boldsymbol{o}_{m}$ and waypoints $\boldsymbol{G}^{\mathcal{B}}_{\mathrm{cur}}$ and $\boldsymbol{G}^{\mathcal{B}}_{\mathrm{nxt}}$ are updated at 10 Hz.
Each episode lasts 14 s and consists of four segments, where each segment has a time budget of $T_{\mathrm{seg}}=3.5$ s. 

\begin{table}[t!]
\captionsetup{font=small}
\caption{Skating-Navigation Policy Rewards $\mathcal{R}=\mathcal{R}_{\text{task}}+\mathcal{R}_{\text{reg}}$}
\label{table:reward}
\vspace{-2mm}
\renewcommand{\arraystretch}{1.4}
\begin{tabularx}{\linewidth}{@{}>{\raggedleft\arraybackslash}p{0.31\linewidth}@{\hspace{3pt}}|@{\hspace{3pt}}>{\arraybackslash}X@{}}
    \Xhline{1\arrayrulewidth}
    \textbf{Reward Term} & \textbf{Expression} \\
    \hline
    \multicolumn{2}{c}{\textit{\textbf{Task Rewards}}: \raisebox{0.1ex}{$\mathcal{R}_{\text{task}} = \sum_{k=1}^{2} r_k$}} \\
    \Xhline{1\arrayrulewidth}
    \text{Waypoint Arrival} ($r_1$) & \raisebox{0.1ex}{$w_1 \; \mathds{1}_{t_{\text{seg}}>T_{\text{seg}}-T_r} / [T_r (1 + \|\boldsymbol{G}^{\mathcal{B}}_{\mathrm{cur}}\|_2^2)]$} \\
    \hline
    \text{Skating Gait} ($r_2$) & \raisebox{0.1ex}{$w_2 \; (\sum_{i=0}^{3} \left[S_W(\phi_i) I_{W_i} + S_F(\phi_i) I_{F_i}\right]) / N_{\mathrm{des}}$} \\
\end{tabularx}

\begin{tabularx}{\linewidth}{@{}>{\raggedleft\arraybackslash}p{0.31\linewidth}@{\hspace{3pt}}|@{\hspace{3pt}}>{\arraybackslash}X@{}}
    \Xhline{1\arrayrulewidth}
    \multicolumn{2}{c}{\hspace{-1.23cm}\textit{\textbf{Regularization Rewards}}: \raisebox{0.1ex}{$\mathcal{R}_{\text{reg}} = \sum_{k=3}^{11} r_k$}} \\
    \Xhline{1\arrayrulewidth}
    \text{Glide} ($r_3$) & \raisebox{0.1ex}{$w_3 \; \sum_{i=0}^{3} I_{W_i} [(v^{\mathcal{B}}_{B,x} - r_W \dot{\psi}_i)^2 + (v^{\mathcal{B}}_{W_i,y})^2]$} \\
    \hline
    \text{Body Stability} ($r_4$) & \raisebox{0.1ex}{$w_4 \; |\theta^{\mathcal{W}}_{B,y}| + w_5 \; (|v^{\mathcal{B}}_{B,z}|^{2} + \|\boldsymbol{\omega}^{\mathcal{B}}_{B,xy}\|_2^{2})$} \\
    \hline
    \text{Smooth Joint} ($r_5$) & \raisebox{0.1ex}{$w_6 \; \|\dot{\boldsymbol{q}}\|_2^2 + w_7 \; \|\ddot{\boldsymbol{q}}\|_2^{2}$} \\
    \hline
    \text{Torque Usage} ($r_6$) & \raisebox{0.1ex}{$w_8 \; \|\boldsymbol{\tau}\|_2^2$} \\
    \hline
    \text{Action Rate} ($r_7$) & \raisebox{0.1ex}{$w_9 \; \|\boldsymbol{a}_t - \boldsymbol{a}_{t-1}\|_2^2$} \\
    \hline
    \text{Smooth Action} ($r_8$) & \raisebox{0.1ex}{$w_{10} \; \|\boldsymbol{a}_t - 2\boldsymbol{a}_{t-1} + \boldsymbol{a}_{t-2}\|_2^2$} \\
    \hline
    \text{Nominal Posture} ($r_9$) & \raisebox{0.1ex}{$w_{11} \; \|\boldsymbol{q} - \boldsymbol{q}_{0}\|_2^{2}$} \\
    \hline
    \text{Action Limit} ($r_{10}$) & \raisebox{0.1ex}{$w_{12} \; \mathds{1}_{\text{action limit}}$} \\
    \hline
    \text{Collision} ($r_{11}$) & \raisebox{0.1ex}{$w_{13} \; \mathds{1}_{\text{collision}}$} \\ 
\end{tabularx}

\begin{tabularx}{\linewidth}{>{\arraybackslash}X|>{\arraybackslash}X}
    \Xhline{1\arrayrulewidth}
    \multicolumn{2}{c}{\hspace{-.4cm}\textit{\textbf{Reward Weights}}} \\
    \Xhline{1\arrayrulewidth} 
    \multicolumn{2}{c}{
        \begin{tabularx}{\linewidth}{
            >{\arraybackslash}X
            >{\arraybackslash}X
            >{\arraybackslash}X
            >{\arraybackslash}X
            >{\arraybackslash}X
            >{\arraybackslash}X
            >{\arraybackslash}X}

         \raisebox{0.1ex}{{\hspace{-2mm}\scriptsize$w_{1}$=$7.0$}} & 
         \raisebox{0.1ex}{{\hspace{-2mm}\scriptsize$w_{2}$=$1.0$}} & 
         \raisebox{0.1ex}{{\hspace{-2mm}\scriptsize$w_{3}$=-$0.1$}} & 
         \raisebox{0.1ex}{{\hspace{-2mm}\scriptsize$w_{4}$=-$5.0$}} & 
         \raisebox{0.1ex}{{\hspace{-2mm}\scriptsize$w_{5}$=-$4.0$}} &
         \raisebox{0.1ex}{{\hspace{-2mm}\scriptsize$w_{6}$=-$1e$-$4$}} &
         \raisebox{0.1ex}{{\hspace{-2mm}\scriptsize$w_{7}$=-$2e$-$7$}} \\

         \raisebox{0.1ex}{{\hspace{-2mm}\scriptsize$w_{8}$=-$2e$-$4$}} & 
         \raisebox{0.1ex}{{\hspace{-2mm}\scriptsize$w_{9}$=-$0.3$}} & 
         \raisebox{0.1ex}{{\hspace{-2mm}\scriptsize$w_{10}$=-$5e$-$2$}} & 
         \raisebox{0.1ex}{{\hspace{-2mm}\scriptsize$w_{11}$=-$4e$-$2$}} & 
         \raisebox{0.1ex}{{\hspace{-2mm}\scriptsize$w_{12}$=-$10$}} &
         \raisebox{0.1ex}{{\hspace{-2mm}\scriptsize$w_{13}$=-$10$}} &
         \\
        \end{tabularx}
    } \\
    \Xhline{1\arrayrulewidth}
\end{tabularx}
\begin{tablenotes}
\setlength{\itemindent}{-0.3cm}
\item[] \textbullet\; $\mathds{1}_{\text{condition}}$ returns 1 if the condition is satisfied and 0 otherwise; \\ e.g., $\mathds{1}_{t_{\text{seg}}>T_{\text{seg}}-T_r}$ is 1 during the final $T_r$ of the current segment, where $t_{\text{seg}}$ denotes the elapsed segment time.
\end{tablenotes}
\vspace{-6mm}
\end{table}

\subsubsection{Reward}\label{sec:Reward}
The reward design is inspired by the goal-reaching formulation~\cite{rudin2022advanced}, with two additional terms for adaptive passive-wheel skating.
The Skating Gait reward encourages the desired contact configuration defined by the skating cycle.
The phase-dependent sign functions \(S_W(\phi_i)=2\mathds{1}_{\Gamma(\phi_i)=\mathrm{Glide}}-1\) and \(S_F(\phi_i)=2\mathds{1}_{\Gamma(\phi_i)=\mathrm{Push}}-1\) assign positive signs to desired wheel and foot contacts and negative signs otherwise. 
The resulting contact score is then normalized by the number of desired contacts \(N_{\mathrm{des}}\).
The Glide reward encourages the wheel motion to remain consistent with the body motion during wheel-ground contact by penalizing longitudinal rolling mismatch and lateral wheel velocity, thereby reducing slip during gliding.
We set the Waypoint Arrival reward window to $T_r=0.3$ s for intermediate segments and $T_r=1.0$ s for the final segment, encouraging pass-through at intermediate waypoints and stopping at the terminal one.
TABLE~\ref{table:reward} details the reward
$\mathcal{R}=\mathcal{R}_{\text{task}}+\mathcal{R}_{\text{reg}}$,
including the individual terms and weights.

\subsubsection{Curriculum}
\label{sec:Curriculum}
We employ a command curriculum that progressively expands the waypoint distance range to facilitate learning under the sparse task reward.
Training begins with a waypoint distance of 2 m along the route, allowing the policy to first acquire skating behavior and reactive obstacle avoidance.
Once the Waypoint Arrival reward reaches 80\% of its maximum, the range expands to 8 m, encouraging faster, longer-range traversal within the same time budget.
For terrain difficulty, we follow the game-inspired curriculum~\cite{rudin2022learning}, which progressively adjusts the terrain level based on traversal performance.

\renewcommand{\arraystretch}{0.9}
\definecolor{c1st}{RGB}{140,214,164}
\definecolor{c2nd}{RGB}{209,237,217}
\definecolor{ourrow}{RGB}{246,246,246}

\section{Experimental Evaluation}
\label{sec:exp}
We conduct experiments in both simulation and real-world environments to evaluate \skatenav~for reliable and energy-efficient navigation of passive-wheel quadrupedal robots.
We design the experiments to answer three key questions: 
(i) Can the proposed framework achieve reliable local navigation in passive-wheel systems? 
(ii) How do the goal-reaching objective and adaptive skating cycle contribute to energy efficiency across varying navigation demands? 
(iii) Can the simulation-trained framework transfer robustly to physical hardware for practical deployment?

\begin{table*}[!t]
  \centering
  \small
  \setlength{\tabcolsep}{5.5pt}
  \renewcommand{\arraystretch}{0.75}
  \captionsetup{font=footnotesize}
  \caption{Navigation performance in simulation environments.
  For details of experimental setups, refer to Sec.~\ref{sec:5-A} and Sec.~\ref{sec:5-B}.
  For w/o Adaptive Freq., we report results at the fixed frequencies yielding the highest SR: $2.0$~\si{\hertz} (\textit{Flat}/\textit{BARN-easy}), $1.8$~\si{\hertz} (\textit{BARN-hard}/\textit{Uphill}),
  and $1.2$~\si{\hertz} (\textit{Downhill}).
  For planners, we select the best cost setting (aggressive, balanced, or safe) by average SR: balanced for DWA; aggressive for MPPI and CEM.}
  \label{tab:navigation_performance}
  \vspace{-1mm}
  \begin{tabular}{@{}l *{12}{c} c@{}}
  \toprule
  & \multicolumn{2}{c}{\textit{Flat}}
  & \multicolumn{2}{c}{\textit{BARN-easy}}
  & \multicolumn{2}{c}{\textit{BARN-hard}}
  & \multicolumn{2}{c}{\textit{Uphill}}
  & \multicolumn{2}{c}{\textit{Downhill}}
  & \multicolumn{2}{c}{\textbf{\textit{Average}}}
  & IT \\
  \cmidrule(lr){2-3}
  \cmidrule(lr){4-5}
  \cmidrule(lr){6-7}
  \cmidrule(lr){8-9}
  \cmidrule(lr){10-11}
  \cmidrule(lr){12-13}

  \textbf{Method}
  & SR\up & CoT\dn
  & SR\up & CoT\dn
  & SR\up & CoT\dn
  & SR\up & CoT\dn
  & SR\up & CoT\dn
  & SR\up & CoT\dn
  & [ms] \\
  \midrule

  \rowcolor{ourrow}
  \rule[-0.5ex]{0pt}{2.6ex}\textbf{Ours}
  & \fst{100.0} & \fst{0.512}
  & \fst{\pz89.3} & \fst{0.539}
  & \fst{\pz79.0} & \fst{0.554}
  & \fst{100.0} & \fst{0.976}
  & \fst{\pz98.4} & \fst{0.216}
  & \fst{\pz93.3} & \fst{0.559}
  & 0.28 \\

  \rule[-0.5ex]{0pt}{2.6ex}
  \hspace{0.8em}$\hookrightarrow$ w/o Adaptive Freq.
  & \fst{100.0} & \snd{0.628}
  & \snd{\pz88.0} & \snd{0.695}
  & \snd{\pz78.0} & \snd{0.653}
  & \fst{100.0} & \snd{1.402}
  & \pz70.0 & \snd{0.348}
  & \snd{\pz87.2} & \snd{0.745}
  & 0.28 \\

  \rule[-0.5ex]{0pt}{2.6ex}
  \hspace{0.8em}$\hookrightarrow$ w/o Skating Cycle
  & \pz83.3 & 1.019
  & \pz75.5 & 1.058
  & \pz72.8 & 1.067
  & \pz96.3 & 1.689
  & \snd{\pz84.8} & 0.428
  & \pz82.5 & 1.052
  & 0.27 \\

  \specialrule{0.4pt}{1.5pt}{1.5pt}

  \rule[-0.5ex]{0pt}{2.6ex}DWA+VT~\cite{fox1997dynamic}
  & \fst{100.0} & 0.608
  & \pz73.1 & 0.657
  & \pz55.5 & 0.683
  & \snd{\pz99.8} & 1.046
  & \pz97.7 & 0.341
  & \snd{\pz85.2} & 0.667
  & 0.42 \\

  \rule[-0.5ex]{0pt}{2.6ex}MPPI+VT~\cite{williams2018information}
  & \pz93.4 & 0.606
  & \pz77.4 & 0.650
  & \snd{\pz70.9} & 0.664
  & \pz86.9 & 1.093
  & \pz94.1 & \snd{0.282}
  & \pz84.5 & 0.659
  & 1.31 \\

  \rule[-0.5ex]{0pt}{2.6ex}CEM+VT~\cite{kobilarov2012cross}
  & \pz98.5 & \snd{0.573}
  & \snd{\pz78.8} & \snd{0.622}
  & \pz70.4 & \snd{0.635}
  & \pz69.6 & \snd{1.007}
  & \snd{\pz98.1} & 0.294
  & \pz83.1 & \snd{0.626}
  & 2.48 \\

  \bottomrule
  \end{tabular}

    \begin{tablenotes}
    \footnotesize
    \item[] \textbullet\;
    {\setlength{\fboxsep}{1pt}%
    \colorbox{c1st}{\textbf{Best}}} and
    {\setlength{\fboxsep}{1pt}%
    \colorbox{c2nd}{second-best}}
    results are highlighted within each comparison group,
    with \textbf{Ours} included in both.
    \end{tablenotes}

  \vstab
  \vspace{-4mm}
\end{table*}

\subsection{Benchmark Environments and Evaluation Metrics} 
\label{sec:5-A}
\begin{figure}[!h]
    \centering
    \vspace{-2mm}
    \includegraphics[width=0.99\columnwidth]{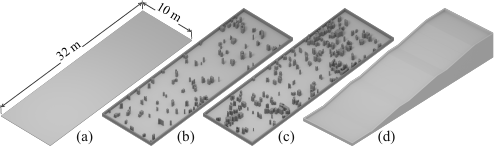}
    \captionsetup{font=footnotesize}
    \caption{
    (a) \textit{Flat}.
    (b)--(c) \textit{BARN-easy} and \textit{BARN-hard}, with obstacle layouts based on
    BARN~\cite{perille2020benchmarking} configurations and mean clearances of $4.26$ and $2.16$~\si{\meter}, respectively.
    (d) The $15^{\circ}$ slope for \textit{Uphill} and \textit{Downhill} conditions.
    }
    \label{fig:terrain}
    \vsfig
\end{figure}

Fig.~\ref{fig:terrain} shows four benchmark environments.
\textit{Flat} evaluates basic traversal ability through skating locomotion. \textit{BARN-easy} and \textit{BARN-hard} assess navigation ability, requiring coordinated propulsion and steering with reactive obstacle avoidance in cluttered environments.
The Slope environment is evaluated under two conditions, \textit{Uphill} and \textit{Downhill}, to assess the ability to sustain propulsion and exploit gliding efficiently, respectively.
For each condition, we generate 1000 navigation tasks by sampling start--goal pairs 8--28~\si{\meter} apart, with four waypoint segments allocated 3.5~\si{\second} each.

We employ three metrics for quantitative evaluation: SR, CoT, and IT.
\textbf{Success Rate (SR)} assesses the proportion of navigation trials in which the robot arrives within \SI{0.2}{\meter} of the target.
\textbf{Cost of Transport (CoT)} evaluates energy efficiency by normalizing the accumulated actuator energy by the robot weight $mg$ and the traveled distance $d$.
The energy proxy is computed using a simplified DC motor model~\cite{seok2014design, shin2023actuator}, in which the electrical power of each actuator is approximated as the sum of mechanical power and Joule loss.
We report the mean CoT over successful trials.
For each trial, CoT is computed as
\vspace{-1mm}
\begin{equation}
    \resizebox{0.90\columnwidth}{!}{$\displaystyle
    \mathrm{CoT} = \frac{1}{mgd}\int_0^T \sum_{j=1}^{12} \max\left(\tau_j\dot{q}_j + \alpha \tau_j^2,\, 0\right) dt, \quad \alpha = \frac{R}{K_t^2}
    $}
    \label{eq:cot}
\end{equation}
Here, $R$=$\SI{0.73}{\ohm}$ and $K_t$=$\SI{1.22}{\newton\meter/\ampere}$ for the actuator~\cite{steadywinGIM8108} used in our system.
\textbf{Inference Time (IT)} evaluates computational efficiency by measuring the time required for each decision step on an RTX 4090 in a single-robot setting.

\subsection{Baselines and Ablation Methods} \label{sec:5-B}
To assess the benefit of coupling navigation decisions with passive-wheel dynamics, we compare \skatenav~(\textbf{Ours}) against planner--controller hierarchies combining a local planner with a Velocity-Tracking (VT) policy.
The VT policy differs from \skatenav~only in the task formulation. 
\textbf{DWA+VT}~\cite{fox1997dynamic} samples dynamically admissible velocity commands and selects the lowest-cost one for the VT policy to track.
\textbf{MPPI+VT}~\cite{williams2018information} samples 512 sequences over a \SI{2}{\second} horizon and applies their softmax-weighted average with an adaptive temperature. 
\textbf{CEM+VT}~\cite{kobilarov2012cross} uses the same horizon and sample budget but iteratively refits a Gaussian proposal to elite samples and applies the converged mean.
All planners share the same rollout model and cost function, comprising goal distance, collision, and command smoothness.

The ablation group comprises two \skatenav~variants designed to evaluate the contributions of the proposed components that shape structured skating behavior: \textbf{w/o Adaptive Freq.}, and \textbf{w/o Skating Cycle}.

\subsection{Navigation Performance}
\label{sec:5-C}
TABLE~\ref{tab:navigation_performance} shows the quantitative results.
\textbf{Ours} achieves the highest Success Rate (SR) and the lowest Cost of Transport (CoT) and Inference Time (IT), outperforming the planner--VT baselines.
For \textit{Flat}, all methods achieve high SR, whereas performance differences become more pronounced as navigation demands increase.
Specifically, \textbf{DWA+VT}~\cite{fox1997dynamic} degrades in \textit{BARN-easy} and \textit{-hard}, where reactive steering corrections for obstacle avoidance are difficult to realize due to intermittent foot-ground propulsion and retained rolling momentum.
Both \textbf{MPPI+VT}~\cite{williams2018information} and \textbf{CEM+VT}~\cite{kobilarov2012cross} mitigate this by optimizing command sequences over a planning horizon, but their navigation decisions remain separated from the phase-dependent skating coordination, reflected in lower SR.

\vspace{-3mm}
\begin{figure}[h]
	\captionsetup{font=footnotesize}
	\centering
	\includegraphics[width=\columnwidth]{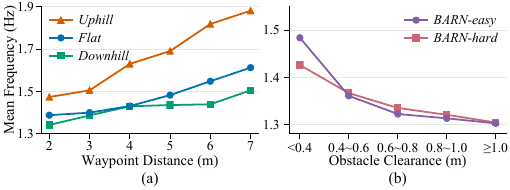}
    \vspace{-6mm}
    \caption{
    Mean skating frequency collected (a) during the first segment of each navigation task across waypoint distances and terrain conditions, and (b) grouped by observed obstacle clearance from the robot center.
	}
	\label{fig:adaptive_freq}
	\vsfig
\end{figure}

In contrast, \textbf{Ours} jointly adapts steering, propulsion, and gliding, with the adaptive skating frequency varying with navigation demand, as shown in Fig.~\ref{fig:adaptive_freq}.
As the waypoint distance increases under the same time budget, the policy raises the skating frequency, producing more frequent \textit{Push} phases.
This tendency is strongest on \textit{Uphill}, where repeated foot-ground contact provides traction for traversal, whereas \textit{Downhill} favors lower frequencies to exploit momentum gained during descent.
Furthermore, the skating frequency increases as obstacle clearance decreases in the \textit{BARN} settings, enabling rapid steering corrections.

\vspace{-3mm}
\begin{figure}[h]
	\captionsetup{font=footnotesize}
	\centering
	\includegraphics[width=\columnwidth]{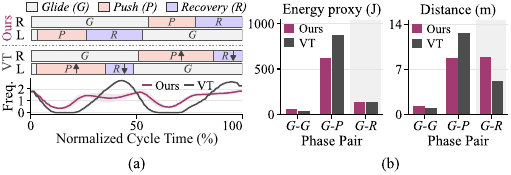}
    \vspace{-6mm}
    \caption{
    Each task comprises four $5$~\si{\meter}, $3.5$~\si{\second} segments, with MPPI+VT commanding $\approx 1.4$~\si{\meter/\second} on average.
    (a) Phase durations and skating frequency over normalized cycle time.
    R/L denote right/left leg pairs; $\uparrow$/$\downarrow$ indicate longer/shorter VT phase durations than Ours.
    (b) Energy proxy and traveled distance accumulated within each phase pair over the evaluation.
	}
	\label{fig:temporal allocation}
	\vsfig
\end{figure}

Beyond adapting to navigation demand, \textbf{Ours} exploits this gait-level flexibility more effectively than VT for energy-efficient skating.
We group all phase pairs between the right and left legs into \textit{Glide--Glide} (\textit{G-G}), \textit{Glide--Push} (\textit{G-P}), and \textit{Glide--Recovery} (\textit{G-R}), regardless of phase order.
As shown in Fig.~\ref{fig:temporal allocation}(a), VT allocates more of each cycle to \textit{G-P} and less to \textit{G-R} than \textbf{Ours}, possibly to track the commanded velocity.
VT consumes more energy during \textit{G-P} than \textbf{Ours} and relies more on this energy-intensive interval for travel, as shown in Fig.~\ref{fig:temporal allocation}(b).
In contrast, \textbf{Ours} travels comparable distances during \textit{G-P} and \textit{G-R} while consuming less energy in \textit{G-R} than in \textit{G-P}.
This temporal redistribution toward momentum-driven motion supports the consistently lower CoT of \textbf{Ours} compared with the planner--VT baselines.

\subsection{Skating Cycle Component Ablation}
\label{sec:5-D}
\begin{figure}[h!]
	\captionsetup{font=footnotesize}
	\centering
	\includegraphics[width=\columnwidth]{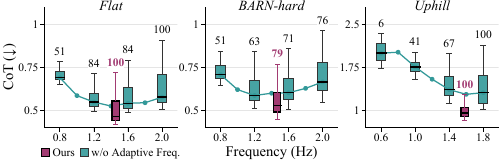}
    \vspace{-4mm}
    \caption{
    Frequency sweep across representative conditions.
    Green boxes show CoT distributions for w/o Adaptive Freq.
    at selected fixed frequencies; lines show medians across the sweep.
    Magenta boxes show \textbf{Ours} at its mean skating frequency.
    Numbers above boxes indicate SR (\%).}
	\label{fig:ablation}
	\vsfig
\end{figure}

\begin{figure*}[t!]
  \captionsetup{font=footnotesize}
  \centering
  \includegraphics[width=\textwidth]{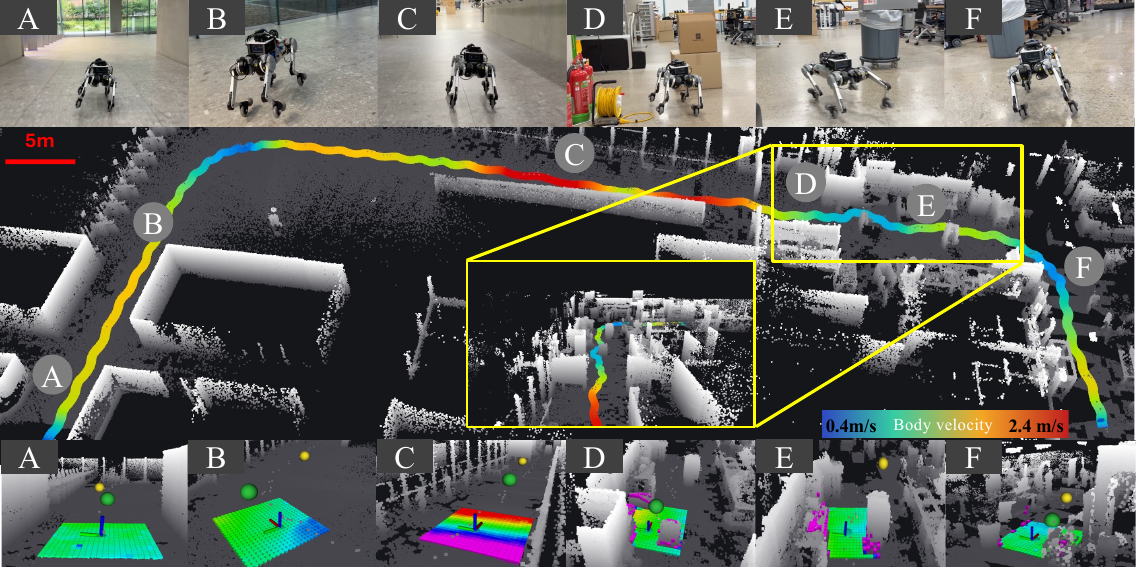}
  \caption{
  \textbf{Real-world navigation example using a passive-wheel quadruped robot.}
  The point-cloud map in the middle row shows the robot trajectory colored by body velocity.
  Markers A--F indicate selected locations along the route, which includes curved passages, a downhill slope, and cluttered indoor spaces.
  The yellow inset provides a top-down view of the marked section.
  The top row shows the robot at each location, and the bottom row shows the corresponding local elevation map provided to the policy, with green and yellow spheres indicating the current and next waypoints, respectively.
  }
  \label{fig:real_world_result}
  \vsfig
\end{figure*}

As shown in Fig.~\ref{fig:ablation}, the skating frequency required for favorable performance varies with navigation demand.
Specifically, \textit{Uphill} requires sufficiently high frequencies for reliable navigation, whereas higher frequencies can raise CoT on \textit{Flat} and \textit{BARN-hard}.
Even within the same environment, different frequency settings lead to noticeable differences in both navigation performance and energy efficiency.
Thus, a single fixed frequency may not achieve both high SR and low CoT across varying navigation demands.
In contrast, \textbf{Ours} learns to adjust the skating frequency as part of its navigation decision, modulating the progression of the skating cycle according to navigation demand. 
Furthermore, without the skating cycle, the goal-reaching objective alone may struggle to discover an energy-efficient skating pattern.
It involves intermittent foot-ground contact to generate propulsion, with the resulting momentum exploited through wheel-ground contact for gliding.
TABLE~\ref{tab:navigation_performance} shows that \textbf{w/o Skating Cycle} yields lower SR and substantially higher CoT across all environments.
This is because the learned motion involves sustained wheel--ground contact accompanied by persistent slip.
The skating cycle therefore provides a useful inductive bias for organizing skating behavior, while adaptive skating frequency provides the flexibility to exploit this structure across varying navigation conditions.

\begin{figure}[h]
  \captionsetup{font=footnotesize}
  \centering
  \includegraphics[width=0.99\columnwidth]{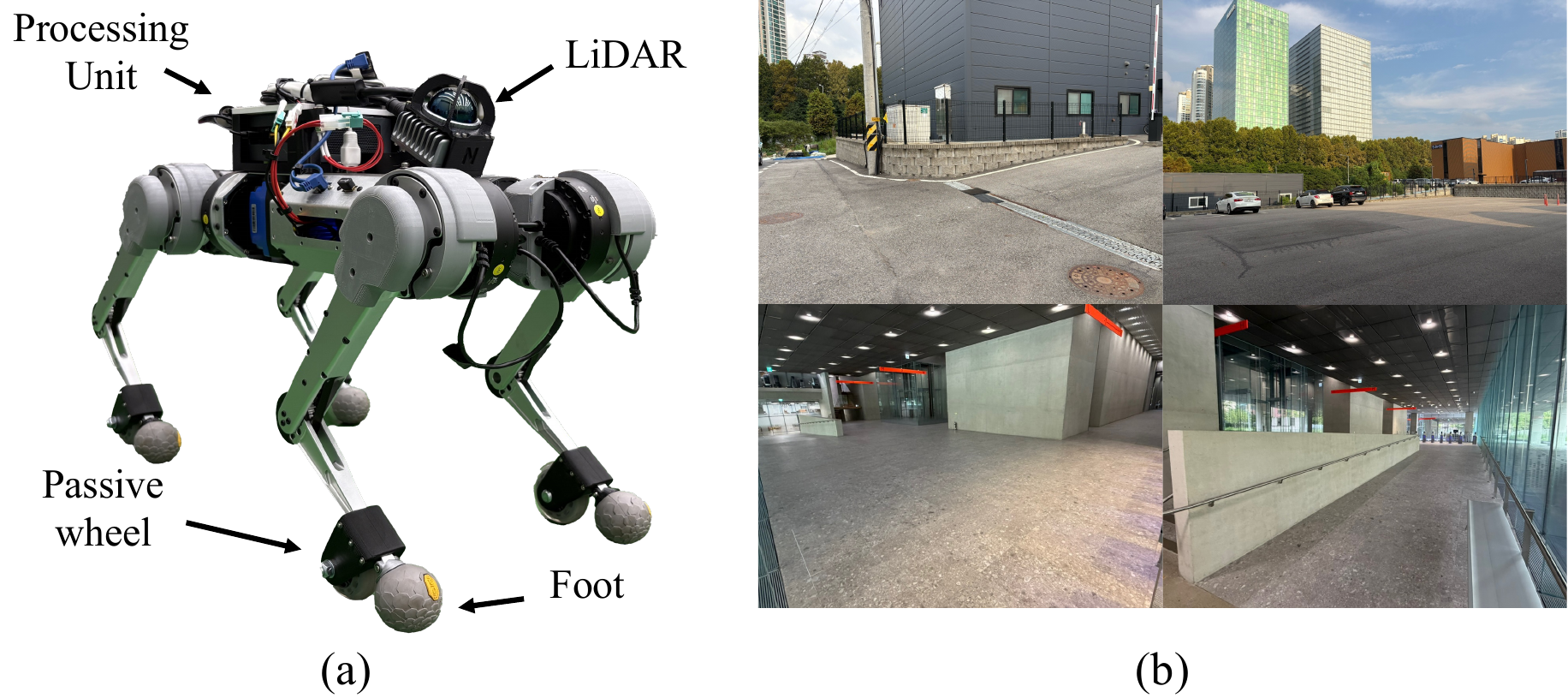}
  \caption{
  (a) \textbf{Hardware setup}. 
  (b) \textbf{Real-world evaluation environments}.}
  \label{fig:real_world_setup}
  \vsfig
\end{figure}

\subsection{Real-World Experiments}
\label{sec:5-E}
We validated \skatenav~on a Mini Cheetah~\cite{katz2019mini} equipped with the passive-wheel modules described in Sec.~\ref{sec:mechanical design} and a Livox Mid-360 LiDAR, using an Intel NUC8-i7 for onboard computation as shown in Fig.~\ref{fig:real_world_setup}(a).
FAST-LIO2~\cite{xu2022fast} provides onboard state estimation, while the terrain perception is obtained using probabilistic elevation mapping~\cite{fankhauser2018probabilistic}.
The elevation map and waypoint commands are updated at \SI{10}{\hertz}, while the policy runs at \SI{50}{\hertz}.

Fig.~\ref{fig:real_world_result} shows a representative indoor navigation scenario along an approximately 57~m route through open and curved areas, cluttered passages, and a downhill section, with 13 commanded waypoints placed 2.5--5.5~m apart.
The scenario requires coordinated steering and propulsion for obstacle avoidance and turning, as well as speed regulation while exploiting momentum downhill.
Despite these challenges, \skatenav~successfully guided the robot to the final goal within the total time budget of 45.5~s, allocated as 3.5~s per waypoint.
The robot slowed to 0.7~m/s through a tight corner, maintained speeds of 0.9--1.2~m/s between obstacles, and accelerated to 1.9~m/s downhill, reaching a peak speed of 2.4~m/s.
We further evaluated \skatenav~in diverse environments, including uphill slopes and paved outdoor surfaces, as shown in Fig.~\ref{fig:real_world_setup}(b).
The supplementary video presents this navigation example and additional experiments.

\section{Conclusion}
\label{sec:conclusion}
In this work, we introduced SkateNav, a unified skating-navigation policy for energy-efficient local navigation of passive-wheel quadrupedal robots.
By combining a time-budgeted goal-reaching formulation with an adaptive skating cycle, \skatenav~jointly coordinates steering and intermittent propulsion while adjusting push--glide timing to navigation demands.
Simulation experiments demonstrated higher average success rates and lower cost of transport than the evaluated planner--controller baselines, while ablation studies confirmed the contributions of the skating cycle and online frequency adaptation.
Real-world experiments further demonstrated zero-shot transfer from simulation across indoor and outdoor environments, including cluttered passages, uphill and downhill slopes, and paved surfaces.

However, the current framework relies on externally specified waypoints and time budgets, which do not explicitly account for the robot's terrain-dependent skating capability.
As future work, we plan to integrate a high-level planner that jointly selects waypoint spacing and time budgets according to terrain conditions and the policy's reachable range, allowing navigation commands to better reflect the propulsion and gliding demands along the planned route.

\bibliographystyle{URL-IEEEtrans}

\bibliography{URL-bib-cleaned}

\end{document}